\documentclass{article}

\usepackage{arxiv}

\usepackage[utf8]{inputenc} 
\usepackage[T1]{fontenc}    
\usepackage{hyperref}       
\usepackage{url}            
\usepackage{booktabs}       
\usepackage{amsfonts}       
\usepackage{nicefrac}       
\usepackage{microtype}      
\usepackage{lipsum}

\usepackage{amsmath}
\usepackage{amssymb}

\usepackage{graphicx}

\title{3D Object Recognition with Ensemble Learning --- A Study of
Point Cloud-Based Deep Learning Models}

\author{
  Daniel Koguciuk
    \\
  Warsaw University of Technology \\
  Faculty of Mechatronics \\
  Boboli 8, 05-525 Warsaw \\
  \texttt{daniel.koguciuk@gmail.com} \\
   \And
  Łukasz Chechliński \\
  Warsaw University of Technology \\
  Faculty of Mechatronics \\
  Boboli 8, 05-525 Warsaw \\
  \texttt{lukasz.chechlinski@gmail.com} \\
   \And
  Tarek El-Gaaly \\
  Voyage \\
  Palo Alto, CA \\
  \texttt{tgaaly@gmail.com} \\
}

\begin{document}
\maketitle

\begin{abstract}
In this study, we present an analysis of model-based ensemble learning for 3D point-cloud object classification and detection. An ensemble of multiple model instances is known to outperform a single model instance, but there is little study of the topic of ensemble learning for 3D point clouds. First, an ensemble of multiple model instances trained on the same part of the \textit{ModelNet40} dataset was tested for seven deep learning, point cloud-based classification algorithms: \textit{PointNet}, \textit{PointNet++}, \textit{SO-Net}, \textit{KCNet}, \textit{DeepSets}, \textit{DGCNN}, and \textit{PointCNN}. Second, the ensemble of different architectures was tested. Results of our experiments show that the tested ensemble learning methods improve over state-of-the-art on the \textit{ModelNet40} dataset, from 92.65\% to 93.64\% for the ensemble of single architecture instances, 94.03\% for two different architectures, and 94.15\% for five different architectures. We show that the ensemble of two models with different architectures can be as effective as the ensemble of 10 models with the same architecture. Third, a study on classic bagging ({\em i.e. with different subsets used for training  multiple model instances}) was tested and sources of ensemble accuracy growth were investigated for best-performing architecture, {\em i.e.} \textit{SO-Net}. We also investigate the ensemble learning of \textit{Frustum PointNet} approach in the task of 3D object detection, increasing the average precision of 3D box detection on the \textit{KITTI} dataset from 63.1\% to 66.5\% using only three model instances. We measure the inference time of all 3D classification architectures on a \textit{Nvidia Jetson TX2}, a common embedded computer for mobile robots, to allude to the use of these models in real-life applications.
\end{abstract}



\section{Introduction}
\label{intro}

Over the last few years with the rapid development of sensor technology, processing of three--dimensional data (3D) has become an important topic of research. High quality, long range laser scanners are widely used in autonomous cars \cite{Geiger2013kitti}, and the availability of cheap RGB-D sensors has resulted in significant progress in 3D mobile robots perception \cite{el2012study}. Accurate object detection, segmentation, and classification from 3D point-clouds are challenging problems, especially so in real-world settings, and crucial for performing robotic tasks.

\begin{figure*}[t]
    \centering
	\includegraphics[width=\textwidth]{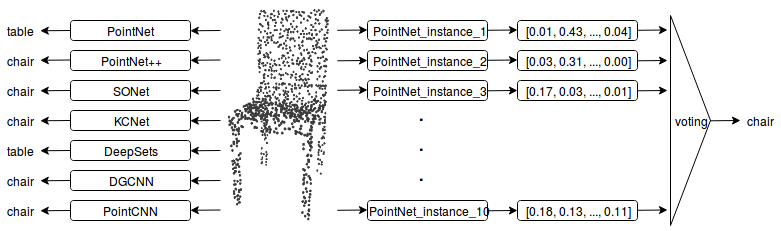}
	\caption{A visualization of an example point cloud from ModelNet40 dataset with classification output of analyzed architectures on the left. On the right output of 10 PointNet instances as a vector of 40 scores for each of class in the ModelNet40 dataset followed by the voting module and with final classification output. All outputs and score values are made up.}
    \label{fig::point_cloud_vis}
\end{figure*}

A lot of the handcrafted approaches to 3D point cloud analysis have been developed previously \cite{himmelsbach2009real, rutzinger2008object, munoz2008directional}; however, in recent years, deep learning -- based approaches have increased in popularity \cite{maturana2015voxnet, sfikas2018ensemble, qi2017pointnet, song2016deep}. Results of deep learning models are strongly correlated to the available amount of data and to the quality of used regularization techniques \cite{goodfellow2016deep}. The study of using these deep neural networks in ensemble learning for 3D point cloud recognition is lacking.

In point cloud classification task we assume that the object is already segmented, which means that all the points belong to that single-class object. There are 3 main approaches to point cloud classification:
\begin{itemize}
    \item 3D ConvNets --- point cloud is converted to voxel grid with a given resolution. This approach is not memory efficient in case of large volumes. The sparsity of 3D data leads to inefficient and redundant computation. However, some octree- or kd-tree--based approaches reduce these disadvantages and provide encouraging results \cite{maturana2015voxnet, Wu2015ShapeNets}.
    \item Rendering a set of 2D views of the 3D object --- the problem is transformed into a set of 2D vision problems. View-pooling layer \cite{su2015multi} may be used to aggregate features from different views. This technique leverages the performance of 3D ConvNets, but the loss of information during rendering makes this approach impractical in point-level segmentation task \cite{sfikas2018ensemble, sarkar2018learning}.
    \item Direct point cloud processing --- architectures that directly process point-clouds in an order-invariant manner, first presented by the \textit{PointNet} architecture. It can be adapted to the analysis of different kinds of problems ranging from an estimation of population statistics \cite{poczos2013regression}, anomaly detection in piezometer data of embankment dams \cite{Jung2015piezometer}, to cosmology \cite{ntampaka2016galaxy}.
\end{itemize}
In this paper, we focus on the direct point cloud processing, because such architectures can perform well not only in classification, but also in segmentation and detection tasks. Seven architectures are used in our experiments: \textit{PointNet} \cite{qi2017pointnet}, \textit{PointNet++} \cite{qi2017pointnet++}, \textit{SO-Net} \cite{li2018sonet}, \textit{KCNet} \cite{shen2018mining}, \textit{DeepSets} \cite{zaheer2017deep}, \textit{DGCNN} \cite{wang2018dgcnn}, and \textit{PointCNN} \cite{li2018pointcnn}. We chose these because of their prominence and the availability of author's implementations that are open to the research community.

For object classification, two types of datasets can be considered. The first type is based on 3D CAD models: \textit{PrincetonSB} \cite{Gan20123DModel}, \textit{ModelNet} \cite{Wu2015ShapeNets}, \textit{ShapeNet} \cite{chang2015shapenet}, and  many others. The second type are datasets of 3D objects/scenes acquired from the real world with depth sensors \cite{Geiger2013kitti, song2015sun, dai2017scannet, Geiger2013kitti}. In this work, we focus on \textit{ModelNet40} \cite{Wu2015ShapeNets}, because it is one of the most popular benchmarks for object classification. It contains 40 classes of objects' CAD models and figure \ref{fig::point_cloud_vis} presents an example point cloud from ModelNet40. For real-world applications, the \textit{KITTI} dataset \cite{Geiger2013kitti} is the most prominent and widely used benchmark for 3D perception of autonomous vehicles and thus is our focus in this study.



\label{fix_C1}
Above mentioned architectures are getting more and more complicated. For example, \textit{PointNet} is a special case of later introduced \textit{PointNet++} with $1.5\%$ increase of instance classification accuracy on \textit{ModelNet40}. However, one can achieve half of this accuracy increase with the ensembling of ten \textit{PointNet} models. We do not want to prove there is no need for further architecture exploration; rather, thanks to ensemble learning we want to gain more insights into those architecture and the task itself.


Ensemble learning \cite{opitz1999popular} increases performance of the prediction by leveraging multiple models. Several methods are reported in the literature: bagging, boosting, stacking, a bucket of models, Bayesian methods, and many others. In this paper, we focus on bagging, also known as bootstrap aggregating. We test three voting methods: direct output averaging, soft voting, and hard voting. We compare the ensemble of model instances trained on the same training set and its different subsets and evaluate their performance.


Our experiments show that an ensemble of neural networks trained on the whole training set is better than bagging using random parts of the training set. An ensemble of \textit{different} model with different architectures can even further improve  performance. In addition, we examine the number of trainable parameters and inference execution times on \textit{NVIDIA Jetson TX2} platform for each approach. According to previous studies \cite{lin2018jetson, tang2017jetson}, using the \textit{Jetson} platform as a high-level driver is a reasonable choice for energy-efficient mobile robotic applications. 



\section{Related Work}
\label{sec::related_work}

To the best of our knowledge, there are no studies reporting on the strict influence of different ensemble methods and the number of aggregated models to the prediction accuracy for direct point cloud classification architectures. Su et al. \cite{su2018deeperlook} studied a model combining different types of representations, but since there are fast advances in this field, there are more and more models in the point cloud classification zoo. A previous article \cite{arvind2017wide} has reported significant performance gain while using an ensemble of 10 instances of one voxel-based, deep learning architecture, introduced in that article.

This section starts with a brief description of each tested architecture. We compare the reported results with test accuracy reproduced in our experiments. The whole setup with exact versions of all libraries and code version used has been shared online\footnote{\url{https://github.com/dkoguciuk/ensemble_learning_for_point_clouds}}. The difference with original reports may be caused by some implementation details, hyper--parameters, or test methodology (reporting maximum vs. average score). Finally, we briefly describe bagging.


\subsection{\textit{DeepSets}}

\textit{DeepSets} \cite{zaheer2017deep} is one of the two independently developed (along with \textit{PointNet}) first deep learning approaches to shape classification using bare point sets. The general idea is quite similar in both methods, but \textit{DeepSets} focuses primarily on producing the permutation equivariance layer, which consists of three operations: $phi$, sum, and $rho$. According to the paper, $phi$ could be an arbitrary neural network architecture applied iteratively over every point in the point cloud, and the output should be summed along the set dimension. The reduced vector can then be passed into the multilayer perceptrion (\textit{MLP}) working as a classifier.

Given the symmetry in weight sharing, the final features of the whole point set (after the summation) are invariant to the ordering of the input. This article demonstrates a strong mathematical background, many sample applications, and great results on the \textit{ModelNet40} dataset. Despite this, there is no direct successor of the model in the literature.

\subsection{\textit{PointNet}}

The basic idea of \textit{PointNet} \cite{qi2017pointnet} is to learn a spacial encoding of each point using a series of nonlinear mappings and then aggregate all point features to one global point cloud signature. The first part plays a similar role to the $phi$ operation in \textit{DeepSets} (there are some slight differences in the weight sharing scheme), and the second one (symmetric function) is also \textit{DeepSets} alike, but the authors suggest using max pooling operation as the one achieving the best performance.

The model is also invariant to the order in which the points are presented, which can directly operate on point clouds without any additional preprocessing such as spatial partitioning or graph construction. Moreover, the model is extremely robust to deformation and noise, but by its design, it suffers from not being able to detect small local structure details; thus, it leaves big space for modifications.

\subsection{\textit{PointNet++}}

\textit{PointNet++} is an extended version of original \textit{PointNet} architecture \cite{qi2017pointnet++}, where authors made hierarchical feature extraction by building a pyramid-like aggregation scheme to combine features from multiple scales. There are three steps on each pyramid level: sampling, grouping, and feature extraction. The first two steps consist of partitioning the input point cloud into overlapping local regions by the distance metric of the underlying space. The third step is learning a higher dimensional representation of an input region with the so-called local learner which is, naturally, a standard \textit{PointNet} model. There are three such pyramid levels in the original \textit{PointNet++} article, which produce the features of the whole point set.

\subsection{\textit{SO-Net}}

\textit{SO-Net} \cite{li2018sonet} is another hierarchical feature extraction model based on \textit{PointNet} approach, but it has a different sampling and grouping strategy in comparison to \textit{PointNet++}. The main idea is to build a Self-Organizing Map (SOM) to model the spatial distribution of the input point cloud and then assign each point into $k$ nearest SOM nodes, which play a similar role to sampling and grouping steps in \textit{PointNet++}. Each local region is processed with a local pointnet-like learner, and a channel-wise max pooling operation is applied to aggregate point features to node signatures. Now, each \textit{SOM} node with its features is processed with the second level learner and again aggregated with max-pool into a feature vector that represents the whole input point cloud.

\subsection{\textit{KCNet}}

The main idea of the \textit{KCNet} model \cite{shen2018mining} is to construct a kernel correlation layer as an affinity measure between a query point with its neighbors and kernel points, where the latter are allowed to move and adjust during training freely. \textit{KCNet} uses several kernels at the local level to augment the original 3D coordinates' input of the \textit{PointNet} architecture. The second modification to the \textit{PointNet} model is a recursive max-pooling operation performed in the neighborhood of each node.

\subsection{\textit{DGCNN}}

Instead of generating the point embedding directly from the point coordinates, the \textit{DGCNN} \cite{wang2018dgcnn} introduces \textit{EdgeConv} operation, which incorporates a point's neighboring structure. For each point of a point cloud, they construct a local neighborhood graph and apply deep learning feature extraction on edges of this graph. \textit{EdgeConv} is designed to be invariant to the ordering of neighbors, and thus is permutation invariant.

The operation could be applied hierarchically just like in the traditional convolutional networks for 2D images, but the authors propose to build the graph of neighboring points for each layer independently.

\subsection{\textit{PointCNN}}

The \textit{PointCNN} \cite{li2018pointcnn} uses a Multi-layer Perceptron (\textit{MLP}) on the local neighborhood to organize points into a latent canonical order. The network learns, so-called, $\mathcal {X} $-transformation results not only in permuting input but also in weighing the features associated with the points. An element-wise product and sum operations are applied on the $\mathcal {X} $-transformed features. These operations can be applied hierarchically: after each convolution, a subset of points are retained by downsampling, thus contain richer information aggregated by the expanding neighborhood.


\subsection{Comparison of Classification Architectures}

\label{fix_C4}
All architectures can be split into the following groups: \textit{DeepSets} and \textit{PointNet} as pionering approaches using global shape feature, \textit{PointNet++} and \textit{SO-Net} as hierarchical pointnets, \textit{KCNet} and \textit{DGCNN} as learnable local feature extractors and \textit{PointCNN} as hierarchical feature extractor.

The general idea behind \textit{PointNet} and \textit{DeepSets} approaches is similar, but they differ mostly in weight-sharing schemes in the \textit{MLP} network and affine transformation matrix prediction by a \textit{T-Net} network in \textit{PointNet}. Authors call it a mini-network, but in fact there are two \textit{T-Net} modules used in \textit{PointNet} (for points and features transformation) and they consist of about $75\%$ of the whole network parameters. Without \textit{T-Net} modules \textit{PointNet} has a similar number of network parameters to \textit{DeepSets}. Both architectures are prominent works in the field but do not explicitly use local structure information.

\textit{PointNet++} and \textit{SO-Net} both apply \textit{PointNet} hierarchically but differ in the sampling and grouping strategy. \textit{PointNet++} samples centroids of local regions using farthest point sampling algorithm (\textit{FPS}) and \textit{SO-Net} uses SOM. Both algorithms reveal the spatial distribution of points, but SO-Net implementation provides deterministic sampling, whereas in \textit{PointNet++} sampled points depend on the choice of the starting point of \textit{FPS}. Secondly, in \textit{SO-Net} each point is assigned into $k$ nearest SOM nodes, which ensure regions to be overlapped. On the other hand grouping in \textit{PointNet++} is done by a ball query within a specified radius -- the radius should be picked carefully so the regions will overlap slightly. The third difference is multi-resolution or multi-scale grouping strategy used in \textit{PointNet++}. \textit{SO-Net} does not have any similar strategy, but experiments with the former show that the strategy does not increase classification accuracy much, rather it helps in robustness against missing points in a point cloud.

\textit{KCNet} and  \textit{DGCNN} both add  local structure information, which is learnable --- not designed by hand like in \textit{PointNet++} or \textit{SO-Net}. \textit{KCNet} uses interesting kernel correlation technique, where \textit{DGCNN} uses feature extraction form graph edges by traditional \textit{MLP}. The former is define only in $\Re^3$, whereas the latter can operate  on high dimensional input, thus can be applied hierarchically. Authors of \textit{KCNet} introduce feature pooling via graphs, which tend to be more effective than max-pool in \textit{PointNet}. \textit{KCNet} and \textit{DGCNN} architectures can be viewed as a \textit{PointNet} working on points with learnable local features and having more effective feature aggregation scheme than original max-pool.

\textit{PointCNN} is the only architecture truly working in hierarchical manner with 4 repeated $\chi$\textit{-conv} operations applied and 3 \textit{FC} layers on top. However, there is a fixed number of neighbors, and they are found using \textit{K-Nearest Neighbor}, which assume an equal distribution of points in the whole point cloud. Besides, there is no clever way of sampling points --- rather they are sampled randomly. Despite all the assumptions \textit{PointCNN} is different from the rest of the architectures because of the permutation invariance approach: it tries to sort points in canonical order rather than using a symmetric function like max-pool.

\subsection{\textit{Frustum PointNet}}
\label{subsec::related_frustum}

The main author of \textit{PointNet} and \textit{PointNet++} extended the work into a 3D object detection framework called \textit{Frustum PointNet} \cite{Qi2018frustum}. The main idea is to combine both 2D and 3D approaches by splitting the task into three main stages:  3D frustum proposal based on 2D object detection, 3D instance segmentation, and 3D bounding box estimation. The modules are based on \textit{PointNet} (denoted as \textit{v1}) or \textit{PointNet++} (\textit{v2}) architectures.

In the first stage, a 2D CNN object detector is used to generate 2D region proposals. Then each region is lifted up to 3D and thus become frustum proposal, containing point cloud $I$ --- all points in the LiDAR point cloud which lie inside the 2D region when projected onto the image plane. The output point cloud $I$  is then fed into 3D instance segmentation network with binary output $P = f_P (\theta_p, I, c)$ (where $\theta_p$ are the model instance weights, $c$ is class of the object and $f_p$ is a model function) meaning if the given point is a part of a 3D object or not, which assumes there is only one meaningful object in the frustum. Points belonging to the object form a point cloud $O = f_o (I, P)$. Then, the T-Net module find transformation $T = f_T (\theta_T, O)$ centering $O$. Centered point cloud is denoted as $C = f_C (O, T)$. In the last step, $C$ is used to estimate amodal 3D bounding box of the object $B = f_B (\theta_B, C)$, which is finally transformed to the global frame. Amodal bounding box $B$ is described by its position, size, and heading angle. Size and heading are represented as discrete probability distributions, not raw values. Summing up, the network prediction can be denoted as:
\begin{equation}
    \begin{split}
        B &= f_B (\theta_B, f_C(O, f_T(\theta_T, O))) \\
        O &= f_O (I, f_P(\theta_P, I, c))
    \end{split}
    \label{eq::frustum_prediction_simple}
\end{equation}

\subsection{Ensemble Learning with Bagging and Boosting}

An ensemble consists of a set of individually trained models, whose predictions are combined. It is well known that ensemble methods can be used to improve prediction performance \cite{opitz1999popular, rokach2010ensemble}.

Individual models may be trained using different training sets. In bagging, the training sets are selected independently for each classifier from the full training set. The selected set can be a subset of the entire training set (later referred to as bagging without replacement) or can have the same size, but samples can repeat (bagging with replacement). However, the result of neural network training depends on several random factors so that the ensemble can consist of classifiers trained on the same training set, which we refer to as a \textit{simple} ensemble.

In Boosting approaches a series of classifiers are tained, with the training set (or samples loss weights) of the next classifier focusing on the samples with a higher error for the previous classifier. This can reduce errors, but noise in the training data often results in boosting overfitting \cite{opitz1999popular}.

Given the output of the individual classifier, the output of the ensemble can be calculated in different ways. Boosting uses individual weight for each of the classifier in series. Stacking trains a learning algorithm to combine predictions. In bagging and \textit{simple} ensembles, all classifiers are equivalent, so three aggregation methods are commonly used in classification task: direct output averaging, soft voting (sum of activation of all hypothesis for each sample equals to one) and hard voting (each classifier output is in one-hot form, {\em, i.e.} each classifier votes for one hypothesis). 



\newcommand{\figureSimpleEnsembleVoting}{
\begin{figure*}[t]
    \centering
	\includegraphics[width=\linewidth]{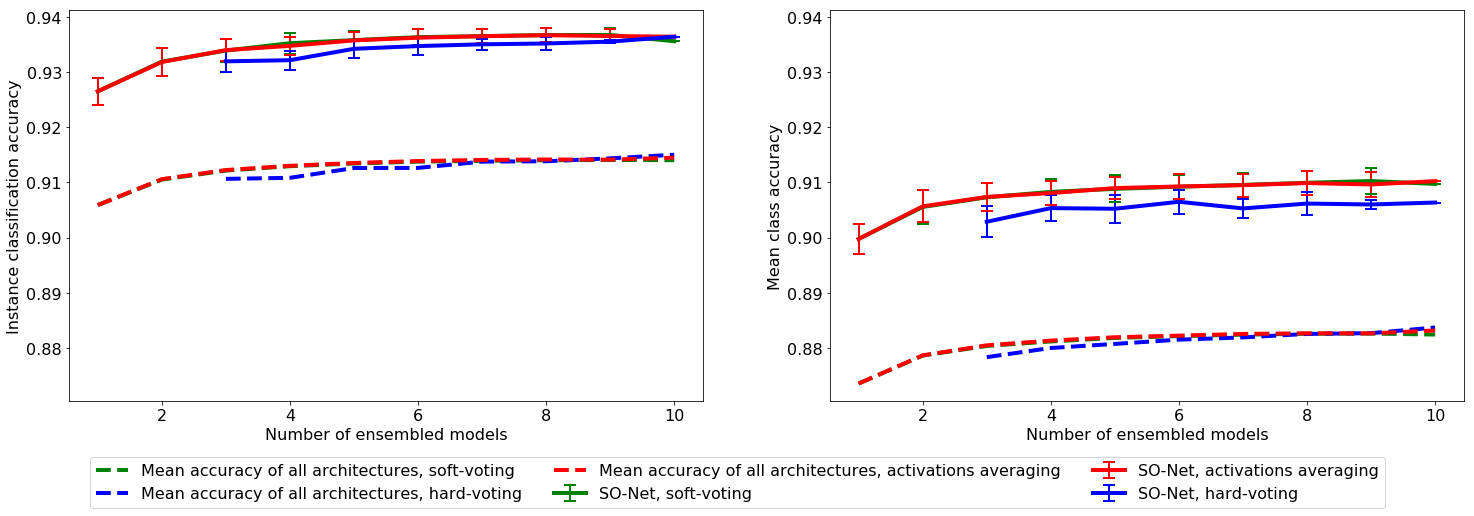}
	\caption{Dependency between instance classification accuracy (left) or mean class accuracy (right) and $k$-number of ensemble models for a given voting method. For visibility, only results for best-performing architecture and mean result of all architectures are plotted. Results for soft-voting and activation ensemble are approximately equal, usually outperforming hard-voting.}
    \label{fig::simple_ensemble_voting}
\end{figure*}}

\newcommand{\figureSimpleEnsembleResults}{
\begin{figure*}[t]
    \centering
	\includegraphics[width=\linewidth]{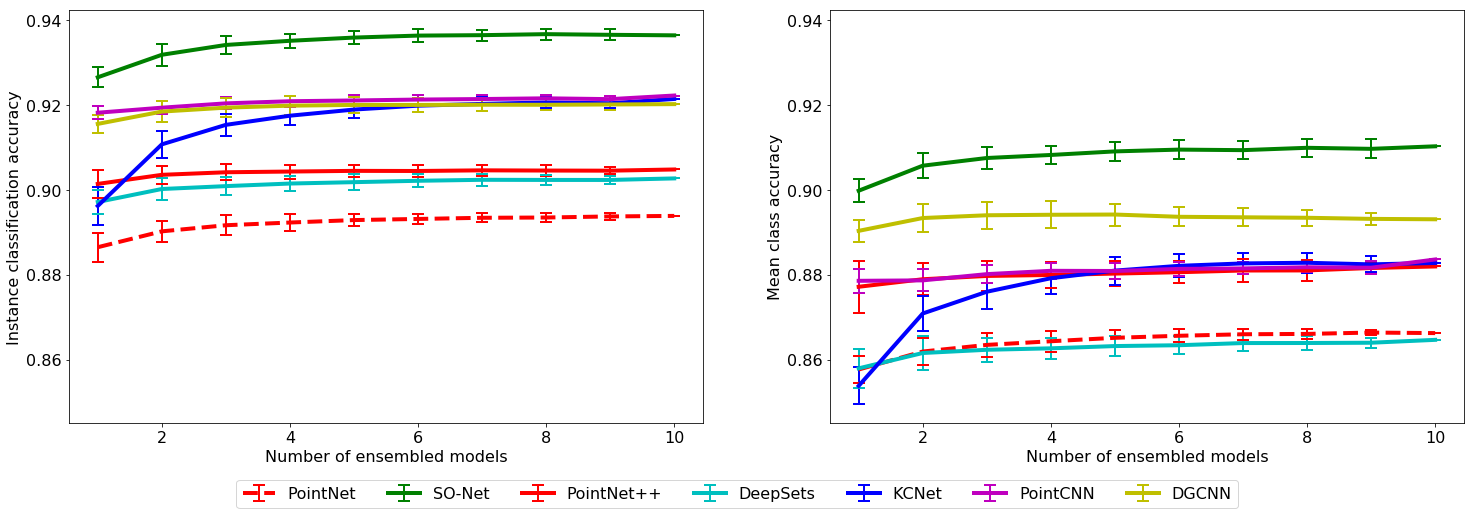}
	\caption{Dependency between instance classification accuracy (left) or mean class accuracy (right) and $k$-number of ensemble models. We have learned each approach independently 10 times (70 different models), then for each $k$ possible number included in the ensemble classifier we have randomly chosen ten different $k$-subsets and have reported mean accuracy with its standard deviation across those $k$-subsets. As one can observe, using the ensemble learning makes the output more stable and classification accuracy rise slightly.}
    \label{fig::simple_ensemble_results}
\end{figure*}}

\newcommand{\figureSimpleEnsembleResultsPerClassLower}{
\begin{figure*}[htbp]
    \centering
	\includegraphics[width=\textwidth]{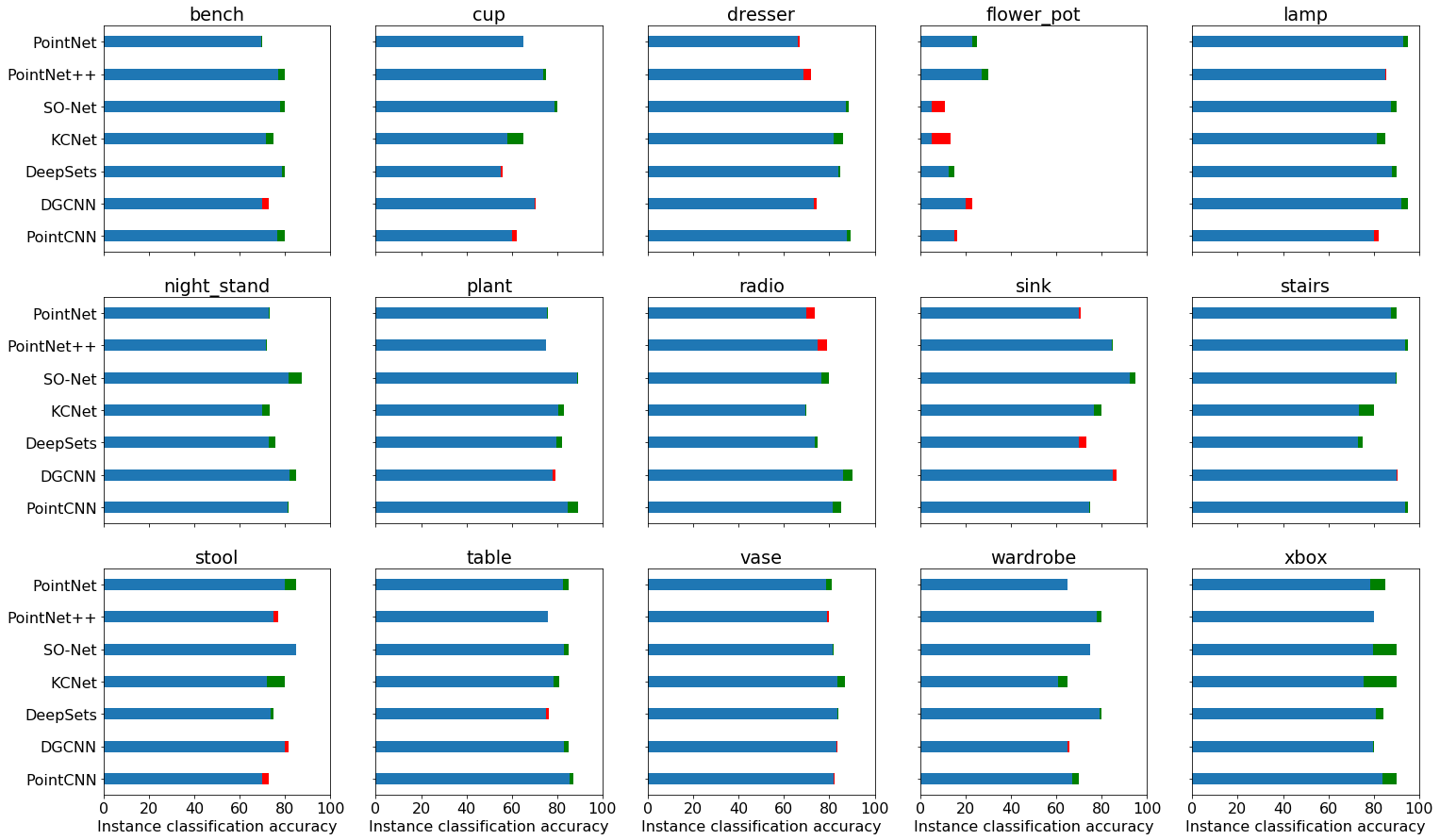}
	\caption{Class classification accuracy, its gain (green), or loss (red) for all architectures for less accurate classes. Please note the \textit{flower pot} class is the hardest class to classify, and almost all architectures using point neighborhood are doing slightly worse with ensemble learning here. This could suggest that the general shape for \textit{flower pot} is more meaningful than local structure information.}
    \label{fig::simple_ensemble_results_per_class_lower}
\end{figure*}}

\newcommand{\figureSimpleEnsembleResultsPerClassHigher}{
\begin{figure*}[htbp]
    \centering
	\includegraphics[width=\textwidth]{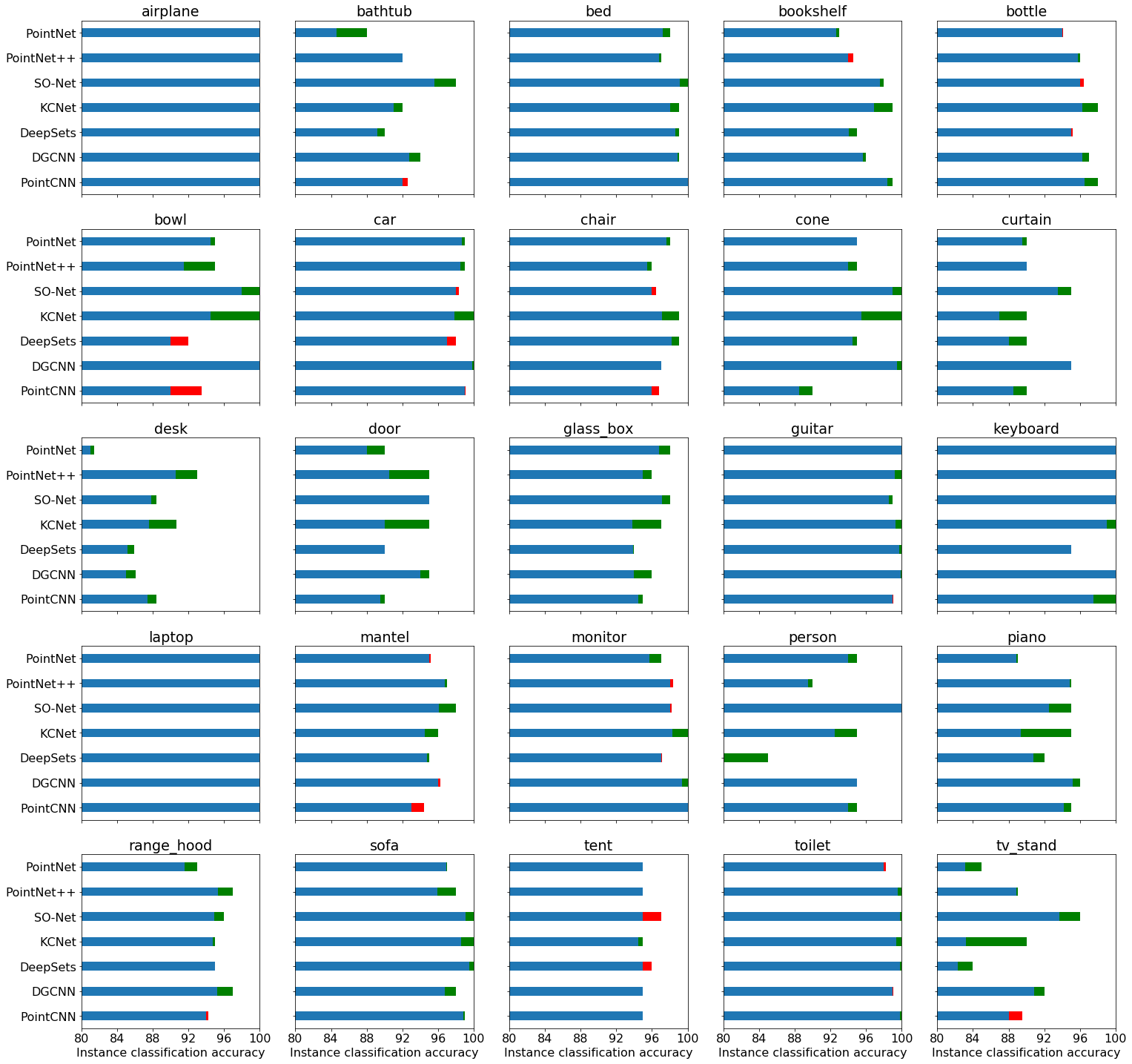}
	\caption{Class classification accuracy, its gain (green) or lose (red) for all architectures for more accurate classes (with accuracy between 80 and 100\%). Please note that many classes, for example, \textit{bed} or \textit{cone}, all architectures achieves better classification accuracy using ensemble learning.}
    \label{fig::simple_ensemble_results_per_class_higher}
\end{figure*}}

\newcommand{\figureSimpleEnsembleWhosTheBest}{
\begin{figure}[htbp]
    \centering
	\includegraphics[width=0.5\textwidth]{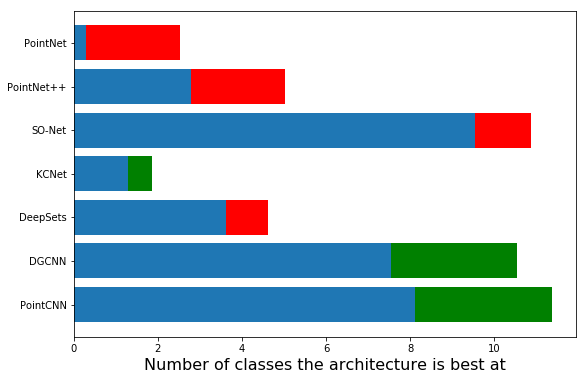}
	\caption{Number of classes where certain architecture is the best for a version without ensemble, its gain (green), or loss (red). High gain of \textit{GDCNN} and \textit{PointCNN} architectures suggests they can be much better in classifying some classes and much worse in other ones.}
    \label{fig::simple_ensemble_whos_the_best}
\end{figure}}


\newcommand{\figureBagging}{
\begin{figure*}[t]
    \centering
	\includegraphics[width=\textwidth]{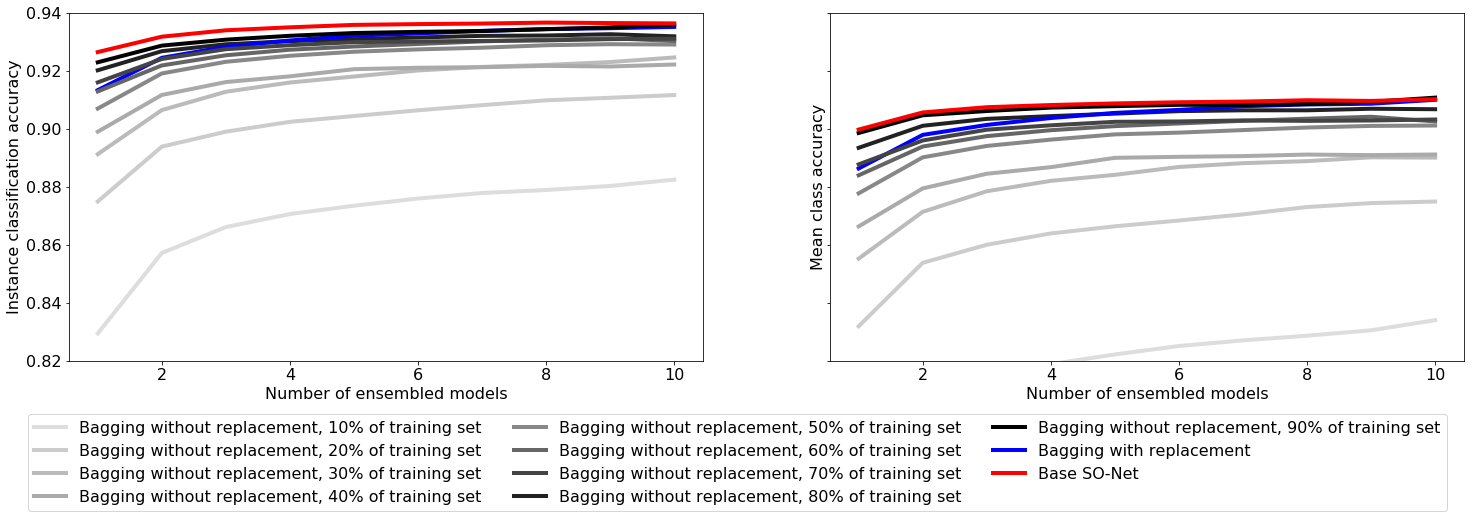}
	\caption{Results of the \textit{simple} ensemble, bagging without replacement for ten different training subset sizes and bagging with replacement. Results for \textit{SO-Net} architecture, instance (left) and class (right) classification accuracy. Error bars not plotted to prevent jamming. The biggest gain is achieved for the smallest training subset size, but the overall classification accuracy is the best for \textit{simple} ensemble (aka bagging with replacement).}
    \label{fig::bagging}
\end{figure*}}


\newcommand{\figureJetson}{
\begin{figure}[t]
    \centering
	\includegraphics[width=0.5\textwidth]{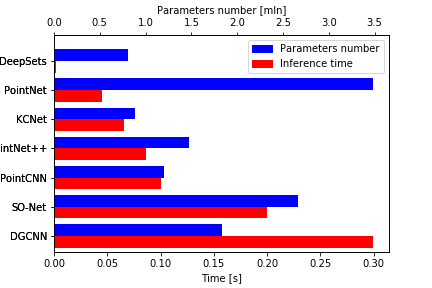}
	\caption{Comparison of the time of inference on the \textit{Jetson TX2} platform (on the bottom) and the number of parameters (on the top) for each tested model. \textit{DeepSets} has the smallest number of parameters and is significantly faster than that of the other approaches.}
    \label{fig::jetson}
\end{figure}}

\newcommand{\tableSimpleEnsembleResults}{
\begin{table*}[b]
\centering
\caption{Average instance and mean class accuracy for each architecture. Results for the single model, \textit{simple} ensemble of 10 models and the accuracy increase are detailed. The \textit{simple} ensemble  of \textit{KCNet} instances  has a high  increase  in  the  classification  accuracy  (2.52\%).}
\resizebox{\textwidth}{!}{
\begin{tabular}{lccccccc}
\hline \hline
           & \begin{tabular}[c]{@{}c@{}}Instance accuracy\\ (reported)\end{tabular} & \begin{tabular}[c]{@{}c@{}}Instance accuracy\\ mean\end{tabular} & \begin{tabular}[c]{@{}c@{}}Instance accuracy\\ ensemble\end{tabular} & \begin{tabular}[c]{@{}c@{}} Class accuracy\\ mean\end{tabular} & \begin{tabular}[c]{@{}c@{}} Class accuracy\\ ensemble\end{tabular} & \begin{tabular}[c]{@{}c@{}} Instance accuracy\\ increase\end{tabular} & \begin{tabular}[c]{@{}c@{}} Class accuracy\\ increase\end{tabular} \\ \hline \hline
\textit{PointNet}     & 89.20     & 88.65     & 89.38     & 85.77       & 86.62     & 0.74      & 0.86     \\
\textit{PointNet++}  & 90.70     & 90.14     & 90.48     & 87.71       & 88.19     & 0.34      & 0.48     \\
\textit{DeepSets}    & 90.30     & 89.71     & 90.27     & 85.79       & 86.46     & 0.56      & 0.67      \\
\textbf{\textit{KCNet}}       & \textbf{91.00}     & \textbf{89.62}     & \textbf{92.14}     & \textbf{85.38}       & \textbf{88.28}     & \textbf{2.52}      & \textbf{2.89}      \\
\textit{SO-Net}      & 93.40     & 92.65     & 93.64     & 89.98       & 91.02     & 0.99      & 1.05  \\ 
\textit{DGCNN}       & 92.20     & 91.55     & 92.02     & 89.0        & 89.30     & 0.47      & 0.27      \\
\textit{PointCNN}    & 92.20     & 91.82     & 92.22     & 87.85       & 88.36     & 0.41      & 0.50      \\
\hline \hline
\end{tabular}}
\label{tab::simple_ensemble_results}
\end{table*}}


\newcommand{\tableDifferentModelsTwoModels}{
\begin{table*}[b]
\centering
\caption{Example results for ensemble of models with two different architectures.}
\resizebox{\textwidth}{!}{
\begin{tabular}{llcccccc}
\hline \hline
 $F_1$ & $F_2$ & $k_1$ & $k_2$      & \begin{tabular}[c]{@{}c@{}}Instance \\ accuracy mean\end{tabular} & \begin{tabular}[c]{@{}c@{}}Instance accuracy \\ ensemble\end{tabular} & \begin{tabular}[c]{@{}c@{}} Class accuracy \\ mean\end{tabular} & \begin{tabular}[c]{@{}c@{}} Class accuracy\\ ensemble\end{tabular} \\ \hline \hline
\textit{SO-Net} & \textit{PointNet}   & 0.7 & 0.3 &   93.23\%   &	93.65\%     & 90.95\%   &	91.44\% 	        \\
\textit{SO-Net} & \textit{PointNet++} & 0.7 & 0.3 &   93.41\%   & 93.75\% 	& 91.33\% 	&   91.61\% 	        \\
\textit{SO-Net} & \textit{KCNet}      & 0.8 & 0.2 &   93.21\%   & 93.73\% 	& 90.62\% 	&   90.99\% 	        \\ 
\textbf{\textit{SO-Net}} & \textbf{\textit{DGCNN}}      & \textbf{0.9} & \textbf{0.1} &   \textbf{93.64\%}   &	\textbf{93.95\%} 	& \textbf{91.59\%} 	&   \textbf{92.00\%} 	        \\ 
\textit{SO-Net} & \textit{PointCNN}   & 0.8 & 0.2 &   93.55\%   & 94.03\% 	& 90.97\% 	&   91.50\% 	        \\ 
\textit{SO-Net} & \textit{SO-Net}     & 0.5 & 0.5 &   93.18\%   &	93.64\% 	& 90.57\% 	&   91.02\% 	        \\ \hline \hline
\hspace{0.1mm}
\end{tabular}}
\label{tab::different_models_two_models}
\end{table*}}

\newcommand{\tableDifferentModelsAllModels}{
\begin{table*}[b]
\centering
\caption{Example results for ensemble of models with different architecture.}
\resizebox{\textwidth}{!}{
\begin{tabular}{cccccccccc}
\hline \hline
$k_{pointnet}$ & $k_{pointnet++}$ & $k_{kcnet}$ & $k_{dgcnn}$ & $k_{pointnet}$ &  $k_{so-net}$ & \begin{tabular}[c]{@{}c@{}}Instance \\accuracy mean\end{tabular} & \begin{tabular}[c]{@{}c@{}}Instance accuracy \\ensemble\end{tabular} & \begin{tabular}[c]{@{}c@{}} Class accuracy \\mean\end{tabular} & \begin{tabular}[c]{@{}c@{}} Class accuracy \\ ensemble\end{tabular} \\ \hline \hline
- & 0.1 & - & - & 0.2 & 0.7 	& 93.73\% & 	94.13\% & 	91.24\% & 	91.77\% \\
0.05 & 0.05 & - & - & 0.25 & 0.65 	& 93.74\% & 	94.14\% & 	91.17\% & 	91.57\% \\
- & 0.05 & - & 0.05 & 0.3 & 0.6 	& 93.88\% & 	94.14\% & 	91.52\% & 	91.92\% \\
- & 0.05 & - & - & 0.2 & 0.75 	& 93.65\% & 	94.15\% & 	91.14\% & 	91.76\% \\
0.05 & 0.05 & - & - & 0.2 & 0.7 	& 93.67\% & 	94.15\% & 	91.12\% & 	91.71\% \\
- & 0.3 & - & 0.05 & - & 0.65 	& 93.73\% & 	94.04\% & 	91.66\% & 	92.20\% \\
0.05 & 0.3 & - & 0.05 & - & 0.6 	& 93.78\% & 	94.04\% & 	91.79\% & 	92.20\% \\
- & 0.15 & - & 0.1 & 0.05 & 0.7 	& 93.76\% & 	94.03\% & 	91.69\% & 	92.21\% \\
- & 0.15 & - & 0.1 & - & 0.75 	& 93.70\% & 	94.05\% & 	91.64\% & 	92.22\% \\
- & 0.2 & - & 0.1 & - & 0.7 	& 93.68\% & 	94.06\% & 	91.60\% & 	92.24\%  \\ \hline \hline
\hspace{0.1mm}
\end{tabular}}
\label{tab::different_models_all_models}
\end{table*}}


\newcommand{\tableLastLayer}{
\begin{table}[t]
\centering
\caption{The influence of classifiers ensemble with one encoder for \textit{SO-Net} architecture. Five classifiers were trained for each of the 10 encoders. As one can see, the computationally cheaper classifier ensemble does not result in rewarding accuracy gain.}
\resizebox{0.5\textwidth}{!}{
\begin{tabular}{lcc}
\hline \hline
Parameter & Mean & Standard Deviation \\ \hline \hline
Instance accuracy mean & 92.43\% & 0.19\% \\
Instance accuracy ensemble & 92.69\% & 0.28\% \\
Class accuracy mean & 89.80\% & 0.21\% \\
Class accuracy ensemble & 89.98\% & 0.28\% \\
Instance accuracy increase & 0.25\% & 0.12\% \\
Class accuracy increase &  0.18\% & 0.12\% \\
\hline \hline
\end{tabular}}
\label{tab::last_layer}
\end{table}}


\newcommand{\tableRandomFactors}{
\begin{table*}[b]
\centering
\caption{The influence of random factors in model instances training to ensemble accuracy gain. One can observe that accuracy increases even if results of massively parallel computations are the only one (irremovable) random factor.
One can observe all removable random factors in the training procedure have little influence on classification accuracy in ensemble learning.}
\resizebox{\textwidth}{!}{
\begin{tabular}{ccccccccc}
\hline \hline
           \begin{tabular}[c]{@{}c@{}}Training \\ data \end{tabular} & \begin{tabular}[c]{@{}c@{}}Weights\\ initialization\end{tabular} & \begin{tabular}[c]{@{}c@{}}Dropout\\ \end{tabular} & \begin{tabular}[c]{@{}c@{}} Instance accuracy\\ mean\end{tabular} & \begin{tabular}[c]{@{}c@{}} Instance accuracy\\ ensemble\end{tabular} & \begin{tabular}[c]{@{}c@{}} Class accuracy\\ mean\end{tabular} & \begin{tabular}[c]{@{}c@{}} Class accuracy\\ ensemble\end{tabular} & \begin{tabular}[c]{@{}c@{}} Instance accuracy\\ increase\end{tabular} & \begin{tabular}[c]{@{}c@{}} Class accuracy\\ increase\end{tabular} \\ \hline \hline
const & const & const & 	92.34\% & 	93.19\% & 	89.58\% & 	90.96\% & 	0.85\% & 	1.38\% \\
const & const & random & 	92.48\% & 	93.11\% & 	89.89\% & 	90.36\% & 	0.63\% & 	0.48\% \\
const & random & const & 	92.50\% & 	93.15\% & 	90.10\% & 	90.58\% & 	0.65\% & 	0.48\% \\
random & const & const & 	92.51\% & 	93.07\% & 	89.88\% & 	90.40\% & 	0.56\% & 	0.52\% \\
random & random & const & 	92.20\% & 	92.99\% & 	89.62\% & 	90.25\% & 	0.79\% & 	0.63\% \\
random & const & random & 	92.47\% & 	93.35\% & 	89.78\% & 	90.51\% & 	0.88\% & 	0.73\% \\
const & random & random & 	92.49\% & 	93.15\% & 	89.84\% & 	90.65\% & 	0.66\% & 	0.81\% \\
random & random & random & 	92.65\% & 	93.57\% & 	89.98\% & 	90.87\% & 	0.92\% & 	0.90\% \\
\hline \hline
\end{tabular}}
\label{tab::random_factors}
\end{table*}}


\newcommand{\tableFrustumClasses}{
\begin{table}[b]
\centering
\caption{\textit{Frustum PointNet} ensemble results for object classes.}
\resizebox{0.5\textwidth}{!}{
\begin{tabular}{lccc}
\hline \hline
Model & Car & Pedestrian & Cyclist \\ \hline \hline
2D proposal & 91.5 & 77.8 & 77.4 \\ \hline
Accuracy ground & & & \\ \hline
\textit{v1}, no ensemble & 81.7 & 62.6 & 63.8  \\
\textit{v1}, last module ensemble & 82.6 & 64.4 & 65.5 \\
\textit{v1}, all modules ensemble & 82.9 & 65.9 & 66.9 \\
\textit{v2}, no ensemble & \textbf{82.3} & \textbf{63.3} & \textbf{65.0} \\
\textit{v2}, last module ensemble & 83.3 & 65.7 & 67.8 \\
\textit{v2}, all modules ensemble & \textbf{83.6} & \textbf{66.5} & \textbf{68.3} \\ \hline
Accuracy 3D & & & \\ \hline
\textit{v1}, no ensemble & \textbf{72.4} & 58.1 & 58.9 \\
\textit{v1}, last module ensemble & 74.1 & 59.4 & 61.5 \\
\textit{v1}, all modules ensemble & \textbf{74.4} & 60.1 & 62.9 \\
\textit{v2}, no ensemble & 72.0 & \textbf{56.5} & \textbf{60.8} \\
\textit{v2}, last module ensemble & 73.6 & 59.6 & 63.7 \\
\textit{v2}, all modules ensemble & 74.0 & \textbf{60.5} & \textbf{65.0} \\ \hline
\hspace{0.1mm}
\end{tabular}}
\label{tab::frustum_classes}
\end{table}}

\newcommand{\tableFrustumDifficulty}{
\begin{table}[b]
\centering
\caption{\textit{Frustum PointNet} ensemble results for difficulty levels.}
\resizebox{0.5\textwidth}{!}{
\begin{tabular}{lcccc}
\hline \hline
Model & Easy & Medium & Hard & All \\ \hline \hline
2D proposal & 90.0 & 79.8 & 76.9 & 82.2 \\ \hline
Accuracy ground & & & \\ \hline
\textit{v1}, no ensemble & 79.1 & 67.6 & 61.5 & 69.4 \\
\textit{v1}, last module ensemble & 80.3 & 69.0 & 63.2 & 70.8 \\
\textit{v1}, all modules ensemble & 81.3 & 70.6 & 63.9 & 71.9 \\
\textit{v2}, no ensemble & \textbf{78.8} & \textbf{68.9} & \textbf{63.0} & \textbf{70.2} \\
\textit{v2}, last module ensemble & 80.8 & 71.2 & 64.8 & 72.3 \\
\textit{v2}, all modules ensemble & \textbf{81.3} & \textbf{71.8} & \textbf{65.3} & \textbf{72.8} \\ \hline
Accuracy 3D & & & \\ \hline
\textit{v1}, no ensemble & 74.3 & 60.5 & 54.5 & 63.1 \\
\textit{v1}, last module ensemble & 76.4 & 62.4 & 56.1 & 65.0\\
\textit{v1}, all modules ensemble & 77.3 & 63.2 & 56.9 & 65.8 \\
\textit{v2}, no ensemble & \textbf{73.5} & \textbf{60.8} & \textbf{54.9} & \textbf{63.1} \\
\textit{v2}, last module ensemble & 76.6 & 63.2 & 57.2 & 65.6 \\
\textit{v2}, all modules ensemble & \textbf{77.5} & \textbf{64.0} & \textbf{58.2} & \textbf{66.5} \\ \hline
\hspace{0.1mm}
\end{tabular}}
\label{tab::frustum_difficulty}
\end{table}}

\section{Ensemble Learning for 3D Object Recognition}

We performed several experiments during the conduct of this study. Seven deep network architectures were selected: \textit{PointNet}, \textit{PointNet++}, \textit{SO-Net}, \textit{KCNet}, \textit{DeepSets}, \textit{DGCNN}, and \textit{PointCNN}. All networks were tested on one task, {\em i.e.} \textit{ModelNet40} object classification. All these networks take a raw 3D point cloud as an input and output a vector of class scores for a given object, which can be denoted as  follows:
\begin{gather}
F: \{p_i \in \Re ^3, i=1, \ldots, N\} \rightarrow \Re ^ C \\
F = \{f_j, j = 1, \ldots, C\} 
\end{gather} 
where $N$ is the number of points in the point cloud, and $C$ is the number of classes in the classification task.



One set of hyper--parameters is selected for each network based on the authors' settings. 10 model instances are trained for each architecture. The influence of the number of models in the \textit{simple} ensemble is tested for each architecture, which is described in subsection \ref{simple_ensemble}.

Different architectures achieve the best result for different object classes. This suggests studying if the ensemble of different models can outperform every single model. Such a comparison for \textit{PointNet, PointNet++, KCNet, DGCNN, PointCNN}, and \ref{sec::different_models}. \textit{DeepSets} was not included due to the challenges in technical implementation.

An \textit{SO-Net} architecture achieved the highest accuracy during our experiments, so it was selected for further bagging tests. We tested the classification accuracy for bagging with and without replacement in subsection \ref{sec::bagging}.

The ensemble of several model instances is computationally expensive. However, a deep network architecture can be viewed as an encoder (transforming sample to a feature vector) followed by a classifier (e.g. \textit{MLP}, calculating class probabilities based on a feature vector). The question arises, whether classification accuracy can be improved by an ensemble of classifiers, based on the same feature vector. This was tested for \textit{SO-Net} architecture according to subsection \ref{sec::last_layer}.

Our work shows that a \textit{simple} ensemble of several model instances of the same architecture increases classification performance. Random factors cause differences between model instances. Influence of each factor is evaluated in  subsection \ref{sec::random_factors}.

An ensemble of three model instances of \textit{Frustum PointNet} was tested on the \textit{KITTI} dataset. This is described in section \ref{frustum}. Note, that \textit{Frustum PointNet} is a pioneering approach, in the sense that it performs detection + classification on raw LiDAR scans, so it cannot be compared directly with the other models presented.



\figureSimpleEnsembleVoting

\subsection{\textit{Simple} Ensemble of Model Instances}
\label{simple_ensemble}

We experiment with the \textit{simple} ensemble, which is a special case of bagging, with the full training dataset being used to train every model instance. The ensembling is performed by averaging the raw output activation, soft voting, or hard voting (denoted as $F$, $S$, and $H$ respectively):
\begin{gather}
    F_e = \{ f_i =  mean(f_{kj}, k = 1, \ldots, K), j = 1, \ldots, C \}
    \\
    S_e = \{ f_i = mean(softmax(f_{k})_j, k = 1, \ldots, K), j = 1, \ldots, C \}
    \\
    H_e = \{ f_i = mean(onehot(argmax(f_{k}))_j, k = 1, \ldots, K), j = 1, \ldots, C \}
    \\
\end{gather}
Where each model instance is denoted as:
\begin{equation}
F_k = \{f_kj, j = 1, \ldots, C\} 
\end{equation}
where $k$ is a model index. For each architecture, 10 model instances were trained. Tests were performed for $K = 1, \ldots, 10$. For each value of $K$, all ${{10} \choose {K}}$ model instances' combinations were selected. Mean and standard deviation for each value of $K$ are reported in the experimental results.

\label{fix_C2}
Figure \ref{fig::simple_ensemble_voting} shows the comparison of voting methods. One needs at least three votes for hard-voting to contribute any useful information. With the infinite number of ensembled models, all voting methods are expected to produce asymptotically same results, but for a finite number of ensembled models, raw activation averaging equals approximately to soft-voting and usually outperforms hard-voting. The difference could be caused by the inflexibility of hard-voting: if a particular model outputs high scores for two classes, a small score change means the instability of the output class. For simplicity, only activation averaging is used in the rest of this paper.


\figureSimpleEnsembleResults

Figure \ref{fig::simple_ensemble_results} presents the instance and mean class accuracy as a function of the number of models in the ensembles for each architecture, along with their standard deviation. As one can observe, with the increasing number of models in the ensemble, the classification accuracy is slightly rising, and the standard deviation of classification accuracy is getting smaller, which means that the output is more stable and not so much dependent on a single learning session.

\tableSimpleEnsembleResults

Table \ref{tab::simple_ensemble_results} shows the numerical comparison of classification accuracy increase between all approaches. The \textit{simple} ensemble of \textit{KCNet} instances has a noticeably higher increase in the classification accuracy (2.52\%), then second \textit{SO-Net} (0.99\%) and other architectures (with mean instance accuracy increase equal to 0.50\%).

\figureSimpleEnsembleResultsPerClassLower
\figureSimpleEnsembleResultsPerClassHigher

Figures  \ref{fig::simple_ensemble_results_per_class_lower} and \ref{fig::simple_ensemble_results_per_class_higher}  show the loss or gain in the classification accuracy per every class between version without and with \textit{simple} ensemble for all tested approaches. Some classes are easy to classify and all approaches achieve 100\% accuracy, for example, airplane and laptop. However, interestingly, there are classes where the \textit{PointNet} approach does better than others, despite the smallest accuracy of overall classification, for example, glass box and stool. This suggests that different methods can be focused on various aspects of point clouds, in particular, focusing on local structure leads to overfitting for some classes with more discriminative global shape.

\figureSimpleEnsembleWhosTheBest

One can ask one more interesting question about those approaches and their ensembles: in how many classes a particular model has the highest accuracy? Figure \ref{fig::simple_ensemble_whos_the_best} answers that question and shows how this number is changed after \textit{simple} ensemble (if $N$ classes reach same best accuracy, each of them scores $1/N$ in this rank). Note that the order of architectures is different than that given in Table \ref{tab::simple_ensemble_results}.


\subsection{Ensemble of Different Models}
\label{sec::different_models}

We evaluate ensembles of pairs of different models. First, an ensemble of output scores were calculated for pairs of architectures according to the following formula:
\begin{equation}
    \begin{split}
        F_{pair} = & k_1 \cdot F_1 + k_2 \cdot F_2 \\
        & k_1 + k_2 = 1 \\
        & k_1 = 0.1, 0.2, \ldots, 0.9
    \end{split}
\end{equation}
Note that each output of the architecture is scaled to have identity standard deviation for the training set. The ensemble of different models improved both instance and mean class accuracy. Top pairs consist of \textit{SO-Net} model with higher weight and the second architecture. Ensemble results are calculated for 10 model instances: 5 of one architecture and 5 of the other. Table \ref{tab::different_models_two_models} shows the results. As two model instances are used in the ensemble to calculate its accuracy, results for two SO-Net instances ensemble are plotted for reference.

\tableDifferentModelsTwoModels

Ensemble of all considered architectures with the best-performing \textit{SO-Net} was tested according to the following formula:
\begin{equation}
    \begin{split}
        F_{all} &= k_{pointnet} \cdot F_{pointnet} + k_{pointnet++} \cdot F_{pointnet++} + k_{kcnet} \cdot F_{kcnet} + \\
        & + k_{dgcnn} \cdot F_{dgcnn} + k_{pointcnn} \cdot F_{pointcnn} + k_{so-net} \cdot F_{so-net} \\
        & k_{pointnet} + k_{pointnet++} + k_{kcnet} + k_{dgcnn}  + k_{pointcnn} + k_{so-net} = 1 \\
        & k_{pointnet} = 0.0, 0.05, \ldots, 0.35 \\
        & k_{pointnet++} = 0.0, 0.05, \ldots, 0.35 \\
        & k_{kcnet} = 0.0, 0.05, \ldots, 0.35 \\
        & k_{dgcnn} = 0.0, 0.05, \ldots, 0.35 \\
        & k_{pointcnn} = 0.0, 0.05, \ldots, 0.35 \\
        & k_{so-net} > 0.4
    \end{split}
\end{equation}
Note that this is the ensemble of different architectures including only one training instance of each architecture. The obtained models are further tested in the aspect of multiple training instances learning as described in subsection \ref{simple_ensemble}. The ensemble of all architectures with the major role of \textit{SO-Net}, achieves the highest overall accuracy. The results of the ensemble are calculated for five instances for each architecture with the nonzero factor. Table \ref{tab::different_models_all_models} shows the most interesting results.

\tableDifferentModelsAllModels


\subsection{Ensemble Learning with Model Bagging}
\label{sec::bagging}

For the \textit{SO-Net} architecture, which achieves the highest overall accuracy, model bagging was tested. For bagging with replacement, 10 training sets were generated by randomly sampling with replacement of a number of samples equal to the size of the training set. For bagging without replacement, $S = 9$ training subset sizes were used in experiments, denoted as follows:
\begin{equation}
    s_i = k \cdot size(T_{train}), k = 0.1, 0.2, \ldots, 0.9
\end{equation}
For each $s_i$, 10 training set splits were generated (sampled without replacement), and one model instance was trained. Output of each model for a given $s_i$ was aggregated as detailed in subsection \ref{simple_ensemble}.

\figureBagging

Figure \ref{fig::bagging} shows the results. The biggest gain is achieved for the smallest training subset size. Accuracy increase for bagging with replacement is higher than that of without replacement. However, none of the bagging methods outperforms \textit{simple} ensemble in the given task.


\subsection{\textit{Simple} Ensemble of Last Layers}
\label{sec::last_layer}

The \textit{SO-Net} architecture consists of the explicitly defined encoder (computationally expensive) and (fast) classifier. Now, we can check whether the accuracy growth of ensemble learning or SO-Net architecture is determined mostly by encoder or classifier part. In the case of the the latter, one could learn the ensemble of classifiers only and thus save learning time by a significant factor.
To check this, for each of the 10 \textit{SO-Net} encoder instances, 5 additional classifiers were trained, with the same hyper--parameters, constant encoder weights, and 31 training epochs. The average result of 5--classifier ensemble is compared to the average result of model with a single classifier (both averages are calculated for 10 encoder instances).

Table \ref{tab::last_layer} shows the results. According to the results, the encoder causes the major advantage of the ensemble. This means that computationally cheaper classifier ensemble does not result in rewarding accuracy gain.


\subsection{Influence of Random Factors in \textit{Simple} Ensembles}
\label{sec::random_factors}

We identified four random factors in \textit{SO-Net} model training:
\begin{itemize}
    \item The order of training samples and random data augmentation (note that eliminating this factor means that samples are shuffled between training epochs but in the same way for every model instance);
    \item Initial values of weights and biases of the neural network;
    \item Random dropout regularization (eliminating this factor means fixing the dropout seed so certain neuron would be dropped, for example, always in epoch number 3, 7, 17, etc.);
    \item Random order of massively parallel computations, resulting in different summation order, which is not alternating for floating point numbers.
\end{itemize}
The first three factors can be eliminated, whereas the last one cannot be eliminated. To verify the influence of each factor, five \textit{SO-Net} model instances were trained for each configuration with one, two, or three random factors eliminated.

\tableLastLayer
\tableRandomFactors

The experiments were time-consuming (35 additional training sessions), but the results, presented in Table \ref{tab::random_factors}, are coarse because only one constant order of values for each factor was considered. However, one can observe that the increase in the accuracy can be observed even if all model instances in the ensemble were trained with the same training data order and augmentation, initial weights, and dropout order. This leads to the conclusion that for \textit{SO-Net} architecture, diversity in models is caused mainly just by the numerical issues of massively parallel computations.


\subsection{Ensemble Methodology for \textit{Frustum PointNet}}
\label{frustum}

Since the first step of \textit{Frustum PointNet} approach use 2D CNN object detector, which is beyond the scope of this paper, we look at ensemble learning for the three point cloud processing modules of the network. A single experiment on the \textit{KITTI} dataset was performed to verify conclusions based on \textit{ModelNet40} and to plan further work on knowledge distillation with ensemble learning for real-world 3D data.

\textit{Frustum PointNet} consists of three trainable modules. Ensembling can be used either for the last bounding-box predicting module or for all modules (denoted as $B_L$ and $B_E$, respectively). Results for other configurations (e.g. ensemble of only segmentation modules) have been shared online. Given $N$ independently trained model instances, $B_L$ and $B_E$ can be defined as follows (compare with equation \ref{eq::frustum_prediction_simple}):
\begin{equation}
    \begin{split}
        B_L &= {1 \over N} \sum_{i=1}^N f_B (\theta_{Bi}, f_C(O, f_T(\theta_T, O))) \\
        B_E &= {1 \over N} \sum_{i=1}^N f_B (\theta_{Bi}, f_C(O_E, {1 \over N} \sum_{j=1}^N f_T(\theta_{Tj}, O_E))) \\
        O_E &= {1 \over N} \sum_{k=1}^N f_O (I, f_P(\theta_{Pk}, I, c))
    \end{split}
\end{equation}

\tableFrustumClasses

\tableFrustumDifficulty

Three \textit{Frustum PointNet} instances were trained for both \textit{v1} and \textit{v2} configurations. Accuracy is calculated for ground and 3D detection. Results are limited by the 2D bounding box proposal accuracy. Objects are divided into three classes (car, pedestrian and cyclist) and three difficulty levels (easy, medium and hard). Tables \ref{tab::frustum_classes} and \ref{tab::frustum_difficulty} presents the accuracy for object classes and for difficulty levels, respectively.

Note that configuration \textit{v1} (based on \textit{PointNet}) often outperforms \textit{v2} (based on \textit{PointNet++}) when ensemble is not used, whereas with ensemble of 3 model instances, \textit{v2} is better. The performance gain is also higher than that for \textit{ModelNet40} classification, which suggests that training set size is smaller concerning the complexity of the task. Averaging outputs of all modules outperforms averaging of only last module output.


\subsection{Comparison of Computational Runtime}
\label{jetson}

We benchmark the speed of all the architectures on the \textit{Jetson TX2} platform. We chose a mini-batch of four point clouds because it seemed to be a reasonable amount of segmented objects visible in a typical mobile robot environment. One has to keep in mind that we have not performed any target-specific optimization. All approaches used (\textit{NVIDIA CUDA}) acceleration and three different deep learning frameworks, based on the original authors' implementations of these methods. 

\figureJetson

\label{fix_C3}
Inference time of deep neural networks depends on many aspects including the number of parameters, depth, number of operations, target-specific optimizations and more. In-depth implementation analysis of considered architectures is beyond the scope of this article, as we want to give a general view on their performance. As depicted in Figure \ref{fig::jetson}, \textit{DeepSets} has the smallest number of parameters and is significantly faster than that of the other approaches, on the other hand \textit{PointNet} has a considerable amount of parameters but is also pretty fast. The results could be explained by the design and types of operations, where \textit{DeepSets} and \textit{PointNet} has simple \textit{MLP-like} structures thus are the fastest, \textit{KCNet} is also pretty fast probably because of small number of parameters, \textit{PointNet++} and \textit{SO-Net} are slower probably because of hierarchical design, and \textit{DGCNN} and \textit{PointCNN} are the slowest since they have graph-based structures, which are expensive to build and convolve over. In the end, it is worth pointing out that an increase in a few percentages of classification accuracy (i.e., \textit{SO-Net} or \textit{PointCNN}) is occupied by significantly longer execution times.






\section{Conclusion}

In this article, we focus on the examination of ensemble learning on 3D point cloud classification with seven most popular architectures using raw point sets. 
We examine the possibility to leverage the classification accuracy of each of the seven cited models by ensemble learning. First, we observe which voting policy is the best for the task. Second, we found slightly better classification accuracy with the increasing number of models in ensemble along with smaller standard deviation. It proves that the ensemble's output is more stable and reliable. The biggest mean instance classification accuracy gain was observed for \textit{KCNet} --- ($2.52\%$), \textit{SO-Net} --- ($0.99\%$), and other for architectures --- ($0.50\%$) on average. Significant increase in classification accuracy achieved by \textit{KCNet} in comparison with all other architectures could be caused by different underlying working principle of kernel correlation as a measure of neuron activation. Or could suggest there is some more space for hyperparameters tweaks in \textit{KCNet}, e.g. the number of filters (sets of kernels).

One can see some interesting observations due to the comparative study of different models. For example, there are classes where the simplest (global) approaches (\textit{PointNet}, \textit{DeepSets}) present the best classification accuracy. This suggests that the general shape of some objects is more important than the shape of local structures, which could mislead more complex models.

We also show that the ensemble models with different architectures can further leverage the overall accuracy. We found that the \textit{SO-Net} got the highest instance and mean class classification accuracy, but \textit{PointCNN} wins in the number of classes in which given network obtained the highest accuracy after ensemble (this score is also high for \textit{DGCNN}). This suggests that the latter can be much better in classifying some classes and much worse in other ones. This could also explain why the ensemble of only two model instances with different architectures lead to approximately $1.\%$ of instance classification accuracy increase compared to the state of the art results. This increase is equal to one obtained while using 10 instances of \textit{SO-Net} model. Further gain can be achieved while using multiple model instances for each of mixed architectures. Instance accuracy of 94.03\% can be obtained for two architectures and 94.15\% while combining three or four architectures.

The performance gain is even higher for \textit{Frustum PointNet'} architecture evaluated on real-world \textit{KITTI} dataset, using only three model instances.

We tested the source of randomness in ensemble learning analysis for \textit{SO-Net}. We observed that numerical issues of massively parallel computations in deep neural networks are essential and beneficial in ensemble learning. The ensemble of several classifiers with the same encoder does not result in a significant performance gain. \textit{Simple} ensemble outperforms classic bagging for tested approaches.

In addition we provide some tips for implementing point cloud classification into a mobile robot equipped with the \textit{Jetson TX2} platform by comparing inference time for all the tested models.

There are more questions one can ask around the topic of ensemble learning for point cloud processing. 
In our opinion, results of this study could leverage the benefit of knowledge distillation in real-world 3D object detection for autonomous cars and mobile robots.

\section*{Acknowledgements}

This research was partially supported by the Dean of Faculty of Mechatronics (Grant No. 504/03731 and Grant No. 504/03272). We want to thank authors of all architectures for providing a public repository. We would also like to gratefully acknowledge the helpful comments and suggestions of Tomasz Trzciński and Robert Sitnik.

\bibliographystyle{unsrt}  
\bibliography{article}

\end{document}